\documentclass{article}

\usepackage{amsmath,amsxtra,amsfonts,amscd,amssymb,bm,amsthm}
\usepackage{graphicx,graphics}
\usepackage{subfig}
\usepackage{harpoon}
\usepackage{geometry}
\usepackage{authblk}

\graphicspath{{figures/}}

\newcommand{\bR}{\mathbb{R}}

\newcommand{\overbar}[1]{\mkern 1.5mu\overline{\mkern-1.5mu#1\mkern-1.5mu}\mkern 1.5mu}
\newcommand*{\vect}[1]{\overrightharp{\ensuremath{#1}}}

\begin{document}

\title{\textbf{Model Reduction with Memory and the Machine Learning of Dynamical Systems}}

\author{Chao Ma\thanks{The Program in Applied and Computational Mathematics, Princeton University, Princeton, NJ 08544, USA}, Jianchun Wang\thanks{Department of Mechanics and Aerospace Engineering, Southern University of Science and Technology, Shenzhen 518055, People’s Republic of China}, Weinan E$^*$\thanks{Beijing Institute of Big Data Research, Beijing, 100871, P.R. China}}

\date{}

\maketitle

\begin{abstract}
The well-known Mori-Zwanzig theory tells us that
model reduction leads to memory effect. For a long time, modeling the memory effect accurately and efficiently has been an important but nearly impossible task in developing a good reduced model.
 In this work, we explore a natural analogy between recurrent neural networks and the Mori-Zwanzig formalism to establish a systematic approach for developing reduced models with memory. 
Two training models-a direct training model and a dynamically  coupled training model-are proposed and compared. We apply these methods to the Kuramoto-Sivashinsky equation and the Navier-Stokes equation. Numerical experiments show that the proposed method can produce reduced model with good performance on both short-term prediction and long-term statistical properties. 
\end{abstract}
\vspace{10mm}

In science and engineering, many high-dimensional dynamical systems are too complicated to solve in detail. Nor is it necessary since
usually we are only interested in a small subset of the variables representing the gross behavior of the system.
Therefore, it is useful to develop reduced models which can approximate the variables of interest without solving the full system. This is the celebrated model reduction problem. 
Even though model reduction has been widely explored in many fields, to this day there is still a lack of systematic and reliable methodologies for model reduction.
One has to rely on uncontrolled approximations  in order to move things forward.

On the other hand, there is in principle a rather solid starting point, the Mori-Zwanzig (M-Z) theory, for performing model reduction \cite{mori1965}, \cite{zwanzig1973}.
In M-Z, the effect of unresolved variables on resolved ones is represented as a memory and a noise term, giving rise to the so-called generalized Langevin equation (GLE).
Solving the GLE accurately is almost equivalent to solving the full system, because the memory kernel and noise terms contain the full information for the
 unresolved variables. 
This means that the M-Z theory does not directly lead to a reduction of complexity or the computational cost. 
However, it does provide a starting point for making approximations. In this regard, we mention in particular the $t$-model proposed by Chorin et al \cite{chorin2002}.
 In \cite{pa17} reduced models of the viscous Burgers equation and $3$-dimensional Navier-Stokes equation were developed by analytically approximating the memory kernel in the GLE using the trapezoidal integration scheme.
Li and E \cite{li2007} developed approximate boundary conditions for molecular dynamics using linear approximation of the
 M-Z formalism.
In \cite{li2017}, auxiliary variables are used to deal with the non-Markovian dynamics of the GLE. 
 Despite all of these efforts, it is fair to say that there is still a lack of systematic and reliable procedure for approximating the GLE. In fact, dealing with the memory terms explicitly does not seem to be a promising approach for deriving systematic and reliable approximations to the GLE.

 One of the most successful approaches for representing memory effects has been the recurrent neural networks (RNN) in machine learning.
Indeed there is a natural analogy between RNN and M-Z. The hidden states in RNN can be viewed as a reduced representation
of the unresolved variables in M-Z.
We can then view RNN as a way of performing dimension reduction in the space of the unresolved variables.
In this paper, we explore the possibility of performing model reduction using RNNs. 
We will limit ourselves to the situation when the original model is in the form of a conservative partial differential equation (PDE), the reduced model is an
averaged version of the original PDE. The crux of the matter is then the accurate representation of the unresolved flux term.

We propose two kinds of models.
In the first kind, the unresolved flux terms in the equation are learned from data. This flux model is then used in
the averaged equation to form the reduced model. We call this the direct training model. A second approach, 
which we call the coupled training model, is to train the neural network together with the averaged equation. 
From the viewpoint of machine learning, the objective in the direct training model is to fit the unresolved flux.
The objective in the coupled training model is to fit the resolved variables (the averaged quantities).

For application, we focus on the Kuramoto-Sivashinsky (K-S) equation and the Navier-Stokes (N-S) equation. The K-S equation writes as 
\vspace{-2mm}
\begin{eqnarray}\label{eq:ks}
&&\frac{\partial u}{\partial t}+\frac 12 \frac{\partial u^2}{\partial x}+\frac{\partial^2 u}{\partial x^2}+\frac{\partial^4 u}{\partial x^4}=0,\ \ x\in\bR, t>0;\\
&&u(x,t)=u(x+L,t),\ \ u(x,0)=g(x).
\end{eqnarray}
We are interested in a low-pass filtered solution of the K-S equation, $\bar{u}$, and want to develop a reduced system for $\bar{u}$. In general, $\bar{u}$ can be written as the convolution of $u$ with a low pass filter $G(y)$:
\begin{equation}\label{eq:avg}
\bar{u}(x,t)=\int_{-\infty}^{\infty}G(y)u(x-y,t)dy.
\end{equation}
Hence, by performing filtering on \eqref{eq:ks}, we get
\begin{equation}\label{eq:ks_avg2}
\frac{\partial \bar{u}}{\partial t}+\frac{1}{2}\frac{\partial \bar{u}^2}{\partial x}+\frac{\partial^2 \bar{u}}{\partial x^2}+\frac{\partial^4 \bar{u}}{\partial x^4}=-\frac{1}{2}\frac{\partial \tau}{\partial x},
\end{equation}
where
\begin{equation}
\tau=\overbar{uu}-\bar{u}\bar{u},
\end{equation}
is the sub-grid stress.
We refer to this as the macro-scale equation, and the original K-S equation as the micro-scale
equation. We refer to $\bar{u}$ as the macro-scale solution.
To develop the reduced system for $\bar{u}$, we need to model the sub-grid stress in \eqref{eq:ks_avg2}. Later we will see that, from the M-Z theory, this stress can be approximated by a memory term of $\bar{u}$.

We also apply the same principle to the $2$-D shear flow, whose governing equation is the following N-S equation
\begin{equation}\label{eq:2dshear}
\partial_t \vect{u}+\left(\vect{u}\cdot D\right)\vect{u}+\nabla p=-\frac{1}{Re}\Delta\vect{u}+\vect{f},\ \ \nabla\cdot\vect{u}=0,
\end{equation}
where $\vect{u}=(u,v)$, $\vect{f}=(f,0)$.
In this problem, the macro-scale solution and the sub-grid stress are defined similarly as for the K-S equation. We 
 model the sub-grid stress using the history of the solution.
Numerical experiments show that our models give good performance on both short-term and long-term predictions.

 Machine learning tools, especially neural networks, have received much attention in recent years.
 Not surprisingly, they have also been used in model reduction such as large-eddy simulation \cite{ga17} \cite{vollant2017} and Reynolds averaged turbulence modeling \cite{ling16}. In these papers, fully-connected neural network models are used to represent the appropriate stress terms. 
 Other machine learnig tools are also used in model reduction.
 For example, in \cite{wang2017}, a random forest based method is employed to improve the Reynolds-averaged Navier-Stokes model. In \cite{xiao2016}, the same problem is studied from a Bayesian viewpoint. 
A time series method is used to construct reduced system for the K-S equation in \cite{Lu16}.
 The difference between our work and these previous papers is that we attempt to develop a systematic approach starting from M-Z, and we explore the natural connection between M-Z and RNNs. In later work, we will further study the mathematical and algorithmic problems along this line.

The paper is organized as follows. Section \ref{sec:mz} briefly introduces the idea of M-Z. Section \ref{sec:rnn} introduces RNNs and LSTMs. In Section \ref{sec:model}, we describe the two training models. Numerical results
on the Kuramoto-Sivashinsky equation and $2$-D shear flow problem 
are present in Section \ref{sec:num}. 
Concluding remarks and comments on future work are presented in  Section \ref{sec:con}.

\section{The Mori-Zwanzig formalism}\label{sec:mz}
The starting point of the M-Z formalism is to divide all the variables into  resolved  and unresolved ones.
The basic idea  is to project functions of all variables into the space of functions of only the resolved variables.
The original dynamical system then becomes  an equation for the resolved variables with memory and noise. This equation is called the Generalized Langevin Equation (GLE). Below we first demonstrate the idea of the M-Z formalism using a simple linear example. 
We then use the Kuramoto-Sivashinsky equation as an example to show how M-Z formalism help us develop a reduced model.

\subsection{A brief review to the M-Z formalism}\label{ssec:mz_1}

 Consider a linear system 
\begin{eqnarray}
\dot{x}=A_{11}x+A_{12}y, \label{eq:linear1} \\
\dot{y}=A_{21}x+A_{22}y, \label{eq:linear2}
\end{eqnarray}
with initial value $x(0)=x_0$, $y(0)=y_0$. 
We take $x$ as the resolved variable,  $y$ as the unresolved variable, and want to develop a reduced system for  $x$. To achieve this, we can first fix $x$ in \eqref{eq:linear2} and solve for $y$, and then insert the result into  \eqref{eq:linear1}. In this way we get
\begin{equation}\label{eq:linear_gle}
\dot{x}=A_{11}x+A_{12}\int_0^t e^{A_{22}(t-s)}A_{21}x(s)ds+A_{12}e^{A_{22}t}y_0.
\end{equation}
  There are three terms at the right hand side of \eqref{eq:linear_gle}. The first term $A_{11}x$ is a Markovian term of $x$; the second term $A_{12}\int_0^t e^{A_{22}(t-s)}A_{21}x(s)ds$ is a memory term.
  The third term $A_{12}e^{A_{22}t}y(0)$ contains the initial value of $y$,  this term will be viewed as a noise term. Hence, by eliminating $y$ from the original linear system, we obtain an equation for $x$ with memory and noise.

Similar deduction can be applied to non-linear ODE systems. Assume we have a system
\begin{equation}
\frac{d\phi}{dt}=R(\phi),\ \ \ \ \phi(0)=x,
\end{equation}
with $\phi$ and $x$ being vectors, and we split $\phi$ into $\phi=(\hat{\phi},\tilde{\phi})$ and take $\hat{\phi}$ as resolved variables, then by the Mori-Zwanzig formalism we can get a GLE for $\hat{\phi}$,
 \begin{equation}\label{eq:gle2}
  \frac{\partial}{\partial t}\hat{\phi}_j(x,t)=R_j(\hat{\phi}(x,t))+\int_0^t K_j(\hat{\phi}(x,t-s),s)ds+F_j(x,t).
  \end{equation}
In \eqref{eq:gle2}, $\hat{\phi}(x,t)$  denotes $\hat{\phi}$ at time $t$ with initial value $x$.
Although \eqref{eq:gle2} is more complicated than \eqref{eq:linear_gle}, the essential ingredients
are similar.
The M-Z formalism tells us that model reduction leads to memory effects, and inspired us to pursuing the application of RNNs as a tool for performing
efficient yet rigorous model reduction.

\subsection{Application to the K-S equation}\label{ssec:mz_2}
 
Going back to the K-S equation \eqref{eq:ks}, let $\hat{u}$ and $\hat{G}$ be Fourier transforms of solution $u$ and filter $G$, respectively. Then, by \eqref{eq:avg}, we have $\hat{\bar{u}}_k=\hat{G}_k\hat{u}_k$ for all frequency $k$. Here, we take a spectrally sharp filter $G$ which satisfies
\begin{equation*}
\hat{G}_k=\left\{ \begin{array}{cc}
1 & if\ |k|\leq K, \\
0 & if\ |k|>K ,
\end{array}\right.
\end{equation*}
for a certain positive integer $K$.
Then, the Fourier transform of $\bar{u}$ is a truncation of the Fourier transform of $u$, and our resolved variables are those $\hat{u}_k$ with $|k|\leq K$.
Writing the K-S equation in Fourier space, and put all terms with unresolved variables to the right hand side of the equation, we can get 
\begin{equation}\label{eq:ks_spec2}
\frac{\partial \hat{u}_k}{\partial t}+(k^4-k^2)\hat{u}_k+\frac{ik}{2}\sum_{\footnotesize\substack{
p+q=k \\ |p|\leq K\\ |q|\leq K}}\hat{u}_p \hat{u}_q = -\frac{ik}{2}\sum_{\footnotesize\substack{p+q=k \\ |p|>K\\ or\ |q|>K}} \hat{u}_p \hat{u}_q.
\end{equation}
Applying Fourier transform to \eqref{eq:ks_avg2}, and compare with \eqref{eq:ks_spec2}, we have
\begin{equation}
\hat{\tau}_k=\sum_{\footnotesize\substack{p+q=k \\ |p|>K\ or\ |q|>K}} \hat{u}_p \hat{u}_q,
\end{equation}
for $|k|\leq K$.
Using the M-Z theory, the sub-grid stress can be expressed as a memory term and a noise term.  
By Galilean invariance, the sub-grid stress can be expressed as a function of the history of 
the resolved strain, $\partial \bar{u}/\partial x$, when the noise effect is negligible.

\section{Recurrent Neural Networks and LSTM}\label{sec:rnn}

In this section, we briefly introduce recurrent neural networks and long short-term memory networks.

\subsection{Recurrent neural networks}
RNNs are neural networks with recurrent connections, designed to deal with time series. Usually, a recurrent connection links some units in a neural network to themselves. This connection means that the values of these units depend on their values at the previous time step. For a simple example, consider  a one-layer RNN with time series $\{x_t\}$ as input. Let $h_t$ and $y_t$ be the hidden values and output values at time step $t$, respectively. Then, the simplest RNN  model is given by
\begin{eqnarray}
h_{t+1}&=&\sigma(Wh_t+Ux_{t+1}+b), \\
y_{t+1}&=&Vh_{t+1}+d, \nonumber
\end{eqnarray}
where $U$, $V$, $W$, $b$, and $d$ are matrices or vectors of trainable parameters, and $\sigma(\cdot)$ is a nonlinear activation function. To deal with complicated time series, sometimes the output of an RNN is fed to another RNN to form a multi-layer RNN.

Note that in RNNs the same group of parameters are shared by all time steps. In this sense, RNNs are stationary models.
Among other things, this allows RNNs to learn information in long time series without inducing too many parameters. 
The training of RNNs is similar to the training of regular feed-forward neural networks, except that we have to consider gradient through time direction, such as the gradient of $y_t$ with respect to $h_{s}$ for $s<t$. Usually RNNs are trained by the so-called back-propagation through time (BPTT) method. 

\subsection{Long short-term memory networks}
Theoretically, RNNs is capable of learning long-term memory effects in the time series. However, in practice it is hard for RNN to catch such dependencies, because of the exploding or shrinking gradient effects \cite{bengio1994}, \cite{pascanu2013}. The Long Short-Term Memory (LSTM) network is designed to solve this problem. Proposed by Hochreiter et al. \cite{hochreiter1997}, the LSTM introduces a new group of hidden units called states, and uses gates to control the information flow through the states. Since the updating rule of the states is incremental instead of compositional as in RNN, the gradients are less likely to explode or shrink. The computational rule of an LSTM cell is
\begin{eqnarray*}\small
f_t & = & \sigma(W_f\cdot[h_{t-1},x_t]+b_f),  \\
i_t & = & \sigma(W_i\cdot[h_{t-1},x_t]+b_i),  \\
o_t & = & \sigma(W_o\cdot[h_{t-1},x_t]+b_o),  \\
\tilde{S}_t & = & \tanh(W_S\cdot[h_{t-1},x_t]+b_S),  \\
S_t & = & (1-f_t)\cdot S_{t-1}+i_t\cdot\tilde{S}_t,  \\
h_t & = & o_t\cdot\tanh(S_t), 
\end{eqnarray*}
where $f_t$, $i_t$, $o_t$ are called the forget gate, input gate, and output gate, respectively, $S_t$ is the state, and $h_t$ is the hidden value. The LSTMs can also be trained by BPTT.

\section{The Two Training Models}\label{sec:model}

For simplicity, we will focus on learning the memory-dependent terms in the GLE, neglecting for now the noise term.
For the K-S equation, as shown in Figure \ref{fig:kstrain}, by directly fitting the stress using the history of the strain, we can reduce the error to nearly $1\%$ while maintaining a small gap between the testing and training error. This shows that we are not overfitting, and the history of the strain determines most of the stress. 

We will discuss two models for learning the memory effect: a direct training model and a coupled training model. 
For both models, we generate data using direct numerical simulation (DNS).
In the direct training model, after generating the data, we directly train the RNN to fit the sub-grid stress $\tau$ using the time history of the strain. 
The loss function is defined as the difference between the output of the RNN and the true stress. The neural network model
for the stress is then used in the macro-scale equation for the reduced system.
In the coupled training model, the loss function is defined as the difference between the solutions of the reduced model (with stress represented
by the neural network) and the ground truth solution.
Therefore in the coupled model, the RNN is coupled with the solver of the macro-scale equation when training.

\subsection{The direct training model}

\begin{figure}[t]
\centering
\includegraphics[width=0.6\textwidth, height=0.4\textwidth]{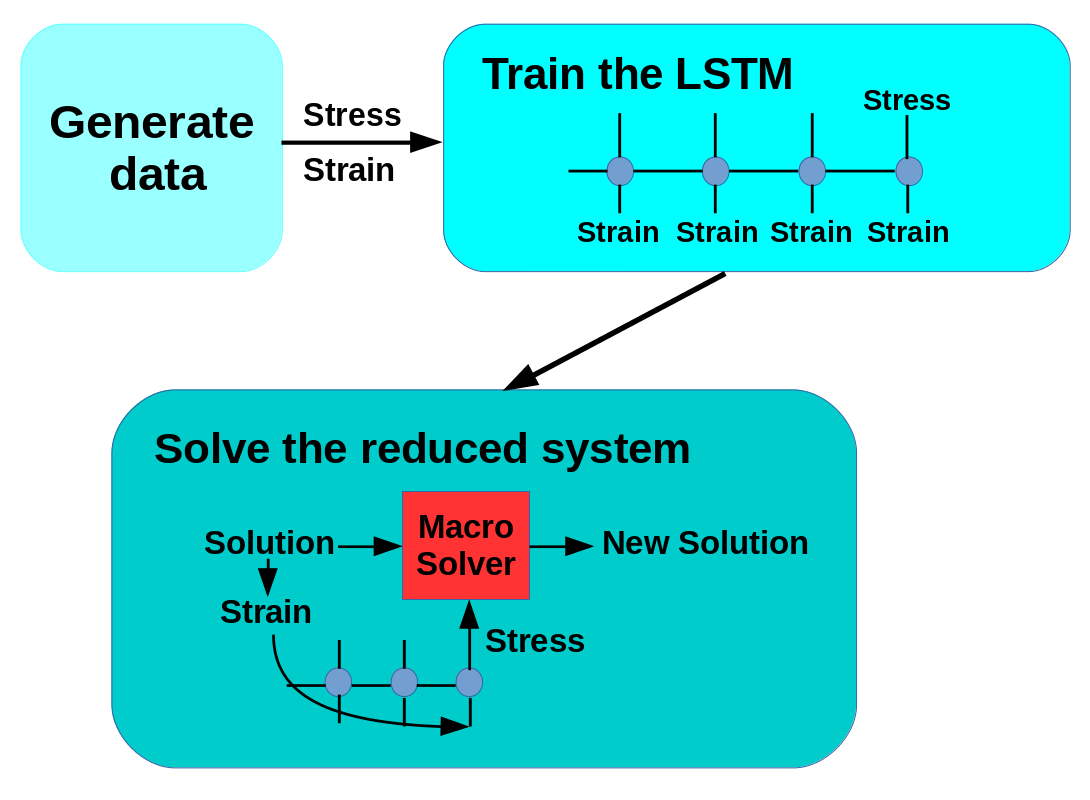}
\caption{The flow chart of the direct training model.}\label{fig:model1}
\end{figure}

In the direct training model, a RNN is used to represent the stress as a function of the time history of the macrocell strain.  
The loss function is simply the squared difference of the predicted stress and the ground truth stress.
This RNN model is then used in the reduced system. 
Figure \ref{fig:model1} shows the flow chart of the direct training model.

\subsection{The coupled training model}\label{ssec:couple}

\begin{figure}[h]
\centering
\includegraphics[width=0.6\textwidth]{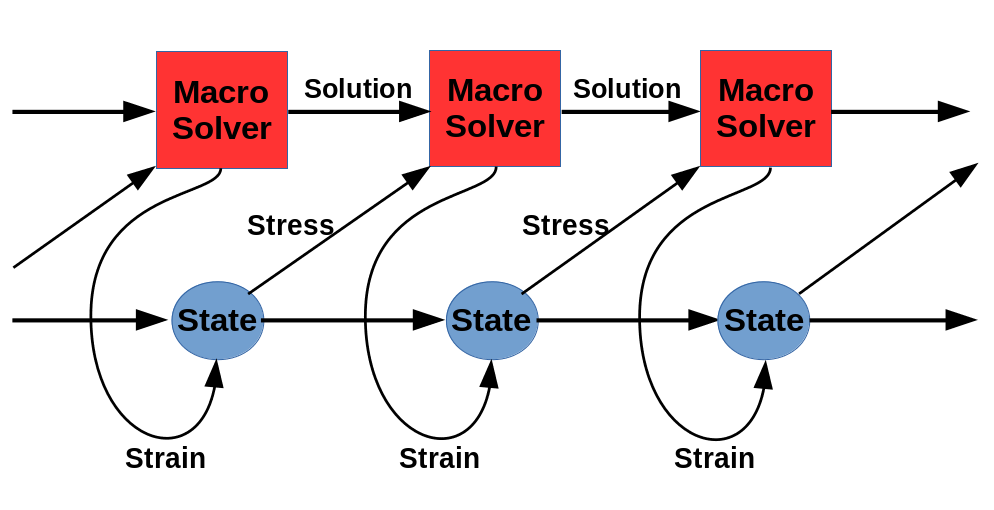}
\caption{The computation graph of the coupled training model.}\label{fig:model2}
\end{figure}

In the direct training model, we train an accurate stress model and we hope that this stress model can help produce accurate results for the macro-scale solution.
In the coupled model,  getting accurate macro-scale solutions is ensured by defining the loss function to be the difference between
the solution of the reduced system and the ground truth, with the stress in the reduced system represented by a RNN.
Specifically, 
 in every epoch, we solve the reduced system for $L$ steps from some initial condition $u_0$ and get a prediction of the solution $L$ steps later ($u_L$). We use $\tilde{u}_L$ to denote the predicted solution. The  loss  function is defined to be some function of $u_L-\tilde{u}_L$. From another perspective, we can also view this coupled system as a single large RNN,  part of the hidden units of this RNN is updated according to a macro-scale solver. Let $\tilde{u}_l$ be the predicted solution at the $l$-th step, and $h_l$, $\tau_l$, $s_l$ be the state of RNN, stress, and strain at the $l-th$ step, respectively. Then the update rule of this large RNN can be written as
\begin{eqnarray}
&&\tau_{l+1}, h_{l+1}=\mbox{RNN}(h_l,s_l), \label{eq:coupled1} \\
&&\tilde{u}_{l+1}, s_{l+1}=\mbox{Solver}(\tilde{u}_l,\tau_l).\label{eq:coupled2}
\end{eqnarray}
Figure \ref{fig:model2} shows part of the computational graph of the coupled training model.

\paragraph{Computation of the gradient}
In the coupled training model, the trainable parameters are still in the RNN, but to compute the gradient of the loss with respect to the parameters, we have to do back-propagation (BP) through the solver of the macro-scale equation, i.e. through \eqref{eq:coupled2}. In many applications, this  solver is complicated and it is hard to do BP through the solver. To deal with this problem, we take one step back to perform BP directly through the differential equations. This is done by writing down the differential equation satisfied by the  gradient, which we call the {\it backward equation}. Another solver is used to solve this backward equation. In this way, BP is done by solving the backward equation, and is decoupled from the macro-scale solver. 

As an illustration, let $\hat{u}(t)$ be the solution of the K-S equation at time $t$ in the Fourier space, and assume that we want to perform back-propagation through the K-S equation from time $t+\delta$ back to $t$, which means we are going to compute the Jacobian $\partial \hat{u}(t+\delta)/\partial \hat{u}(t)$. Let 
\begin{equation}
J(s)=\frac{\partial \hat{u}(t+s)}{\partial \hat{u}(t)},
\end{equation}
then we have $J(0)=I$ and we want to compute $J(\delta)$. In the reduced system, we solve the K-S equation with stress,
\begin{equation}\label{eq:ks_reduced_spec}
\frac{d\hat{u}}{dt}=\mbox{diag}(k^2-k^4)\hat{u}-\frac{i}{2}\mbox{diag}(k)\hat{u}\ast \hat{u}-\frac{i}{2}\mbox{diag}(k)\hat{\tau},
\end{equation}
where $\hat{\tau}$ is the sub-grid stress in Fourier space, $\ast$ means convolution, and $\mbox{diag}(v)$ represents a diagonal matrix with vector $v$ being the diagonal entries. Taking derivative of $J$  with respect to $s$, and assuming that  $\partial \hat{\tau}(t)/\partial \hat{u}(t)=0$, we obtain
\begin{equation}\label{eq:bp_j}
\frac{d J(s)}{ds}=\mbox{diag}(k^2-k^4)J(s)-ik\hat{u}(t+s)\ast J(s).
\end{equation}
Hence, as long as we know the solution $\hat{u}(t+s)$ for $0\leq s\leq\delta$,  we can compute the Jacobian $J(\delta)$ by solving the equation \eqref{eq:bp_j} from $0$ to $\delta$, with initial condition $J(0)=I$.

\section{Numerical Experiments}\label{sec:num}

We now present some numerical results of the proposed models for the K-S equation  
and the 2-dimensional shear flow problem. Below when we talk about true solution or ground truth,  we mean
the exact solution after filtering.

\subsection{The Kuramoto-Sivashinsky equation}\label{ssec:ks}
\paragraph{Experiment setting} 

The K-S equation is considered in the Fourier space. To generate the training data, we solve the K-S equation with $N=256$ Fourier modes to approximate the accurate solution,  using a $3^{rd}$ order integral factor Runge-Kutta method \cite{shu1988}. We set $L=2\pi/\sqrt{0.085}$ and the time step $dt=0.001$. The settings are similar to that in \cite{Lu16}. The micro-scale equation is solved for $1.2\times10^5$ time units and filterd outputs are saved for every $0.1$ time units.
We drop the results of the first $10^4$ time units, and take the results from the following $10^5$ time units as the training data, and the results of the last $10^4$ time units as the testing data.

For the reduced system, we solve \eqref{eq:ks_avg2} in Fourier space.
We take $K=16$ Fourier modes and the time step $dt_r=0.1$. 
The macro-scale solver is still a $3^{rd}$ order integral factor Runge-Kutta method. The solver works in the Fourier space, and takes the output of the RNN as the sub-grid stress.

On the machine learning side, we use an LSTM to predict the stress. 
The LSTM has two layers and $64$ hidden units in each layer. An output layer with linear activation is applied to ensure that the dimension of the outputs is $16$. The LSTM works in the physical space: it takes strains in the physical space as inputs, and outputs predicted stresses in the physical space. A Fourier transform is applied to the output of the LSTM before feeding it into the macro solver.

\begin{figure}[h]
\centering
\includegraphics[width=0.8\textwidth, height=0.35\textwidth]{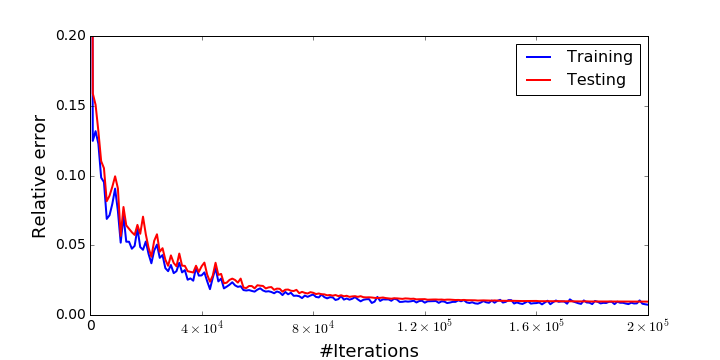}
\caption{Relative training and testing error during training.}\label{fig:kstrain}
\end{figure}

\paragraph{Direct training}
For the direct training model, we train the LSTM to fit the stress as a (memory-dependent) function of the  strains for the last $20$ time steps. The network is trained by the Adam algorithm \cite{kingma2014} for $2\times10^5$ iterations, with batch size being $64$. Figure \ref{fig:kstrain} shows the  relative training and testing error during the training process. We can see that the two curves go down simultaneously, finally reaching a relative error of about $1\%$. There is almost no gap between the training and the testing error, hence little overfitting. This also suggests that most contribution to the sub-grid stress can be explained by
the time history of the strain.

\begin{figure}
\centering
\includegraphics[width=0.4\textwidth]{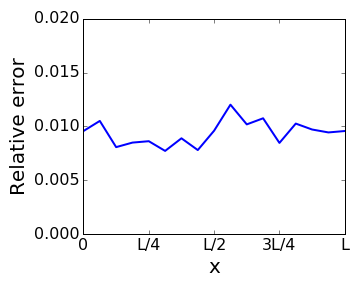}
\includegraphics[width=0.4\textwidth]{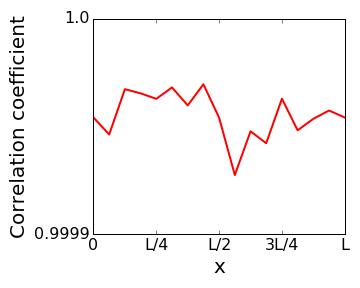}
\caption{Relative error and correlation coefficients of the prediction at different locations in the physical space.}\label{fig:kscorr}
\end{figure}

First we consider some a priori results. Figure \ref{fig:kscorr} presents the relative error and the correlation coefficients between the predicted stress and the true stress at different locations in the physical space. We can see that our prediction of the stress has relative errors of about $1\%$ and correlation coefficients very close to $1$.

\begin{figure}[ht]
\centering
\includegraphics[width=0.4\textwidth]{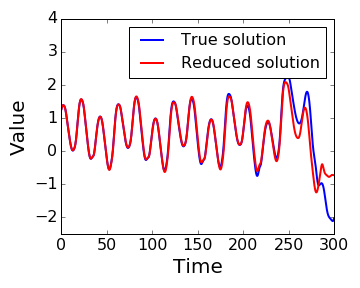}
\includegraphics[width=0.4\textwidth]{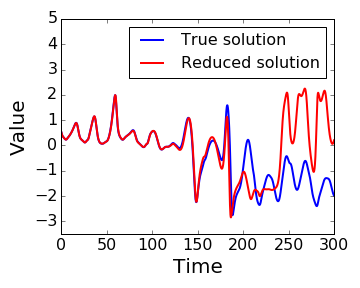}
\caption{Comparison of the reduced solution given by the direct training models and the true solution. The lines shows values of the solutions at a certain location in the physical space.}
\label{fig:diphy}
\end{figure}

We next examine some a posteriori results. After training the model, we pick some times in the testing set, and solve the reduced system initialized from the true solution at these times. Then, we compare the reduced solution with the true solution at later times. 
Figure \ref{fig:diphy} shows two examples. In these figures, the $x$-axis measures the number of time units from the initial solution. From these figures we can see that the reduced solution given by the direct training model produces satisfactory prediction for $150$-$200$ time units ($1500$-$2000$ time steps of macro-scale solver). 

As for the eventual deviation between the two solutions, it is not clear at this point what is more responsible, the intrinsic unstable behavior in the model or the model error. We will conduct careful studies to resolve this issue in future work.

\begin{figure}
\centering
\includegraphics[width=0.4\textwidth]{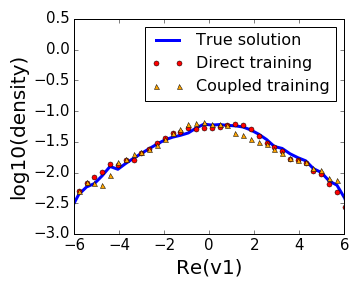}
\includegraphics[width=0.4\textwidth]{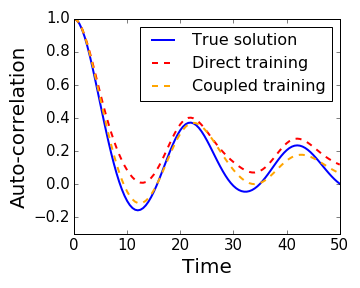}
\caption{ Distribution and auto-correlation of the real part of the Fourier mode with wave number $k=1$. The direct training model and the coupled training model.}\label{fig:didist}
\end{figure}

Next we compare the long-term statistical properties of the solution to the reduced model with the true solution. We solve the reduced system for a long time ($10^4$ time units in our case).  We then compute the distributions and the autocorrelation curves of Fourier modes of the reduced solution, and compare with that of the true solution. Figure \ref{fig:didist} shows some results. From these figures we see that the reduced model can 
satisfactorily recover statistical properties of the true solution.  

\begin{figure}
\centering
\includegraphics[width=0.75\textwidth]{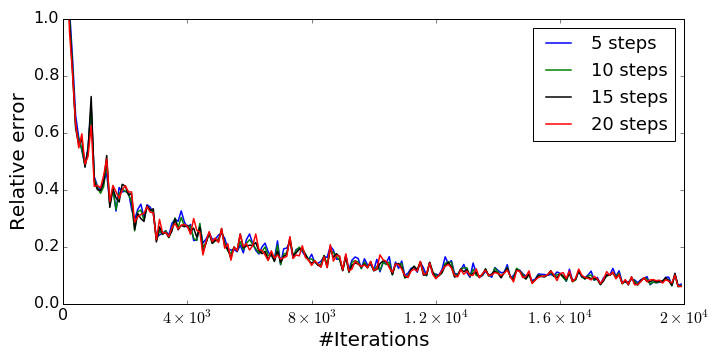}
\caption{Relative error at different number of steps from the initial true solution during the training process.}\label{fig:train}
\end{figure}

\paragraph{Coupled training}
For the coupled training model, we train the LSTM together with the equation solver. A detailed description of this training model is given in Section \ref{ssec:couple}. Here we choose $L=20$ during training. The size of the LSTM is the same as that in the direct training model. A simple one-step forward scheme is used to solve the backward equation for the Jacobian \eqref{eq:bp_j}.  $2\times10^4$ iterations were performed using an Adam optimizer with batch size $16$. During the training process, we measured the relative error of predicted solutions $l$ steps later from the initial true solution. The results for different $l$'s are given in Figure \ref{fig:train}. From the figure we can see that the relative error for different $l$ goes down to $6-7\%$ after training, and there is no gap between the different curves, which means that solutions with different steps from the initial time are fitted nearly equally well.

\begin{figure}[t]
\centering
\includegraphics[width=0.4\textwidth]{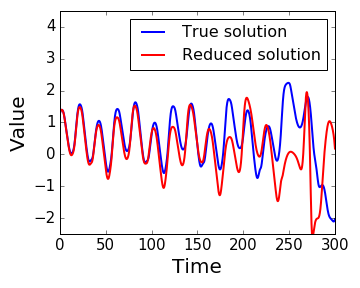}
\includegraphics[width=0.4\textwidth]{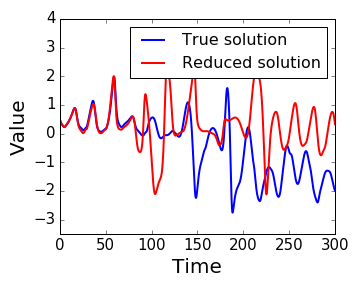}
\caption{Comparison of the reduced solution and the true solution on the physical space. }\label{fig:cpphy}
\end{figure}

Short-term prediction and long-term statistical performance are also considered for the coupled training model. Results are shown on Figure \ref{fig:cpphy} and \ref{fig:didist}. From the figures we see that, the reduced system trained by the coupled training model gives 
satisfactory prediction for $50-100$ time units, which is shorter than the direct training model. However, the auto-correlation of Fourier modes match better with the true solution, while the distribution is as good as the direct training model. This suggests that, compared to the direct training model, the coupled 
model is less competitive for short-term prediction, but performs better for  long-term statistical properties.

\subsection{The $2$-dimensional shear flow}\label{ssec:2d}
\paragraph{Experiment setting}

Consider the $2$-dimensional shear flow in channel, whose governing equation is given by \eqref{eq:2dshear}.
We take $(x,y)\in[0,100]\times[-1,1]$ and use periodic boundary condition in $x$ direction and zero boundary condition at $y=\pm1$. We choose $Re=10000$, and $f=2/Re$ as a constant driving force. To numerically solve the equation, we employ the spectral method used in \cite{weinan2016}. We take $256$ Fourier modes in $x$ direction and $33$ Legendre modes in $y$ direction. The time step $\Delta t$ are chosen to be $0.005$.

For ease of implementation, we only do model reduction for Fourier modes in $x$ direction, and keep all Legendre modes in $y$ direction. The macro solution has $64$ Fourier modes in $x$ direction. Still, we generate accurate solution by DNS and compute the macro-scale strain and stress, and use an LSTM to fit the stress as a function of the history of the strain. In this experiment, the neural network we use is a $4$-layer LSTM with $256$ hidden units in each layer. As for the K-S equation, the input and output of the LSTM are in the physical space.

\begin{figure}
\centering
\includegraphics[width=0.8\textwidth]{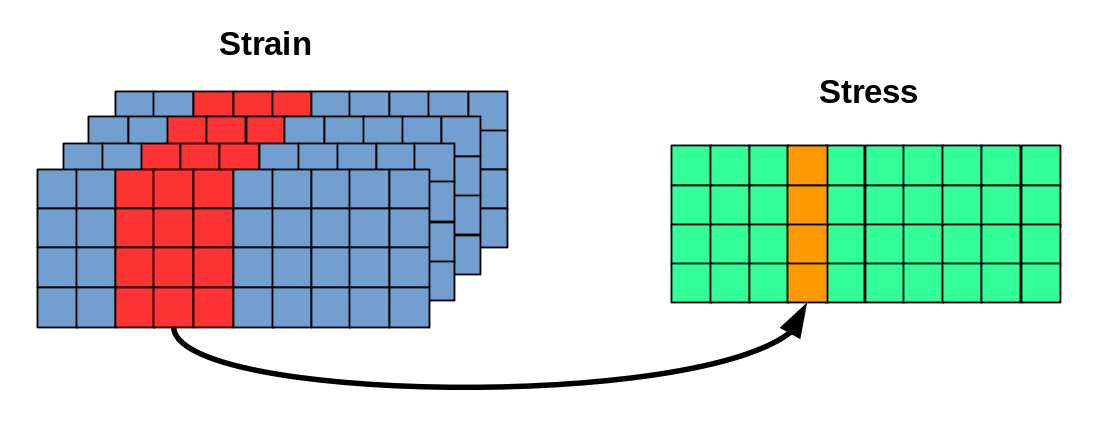}
\caption{The LSTM predicts one column of the stress each time, using strains at this column and the neighboring columns.}\label{fig:fitcol}
\end{figure}

In this problem, the stress has $3$ components ($\tau_{11}=\overbar{uu}-\bar{u}\bar{u}$, $\tau_{12}=\overbar{uv}-\bar{u}\bar{v}$, $\tau_{22}=\overbar{vv}-\bar{v}\bar{v}$). Since each component has $33\times64=2112$ modes in the spectral space, we need at least $2112$ variables in the physical space to represent the stress. If directly fit the stress, our LSTM will have about $6$ thousand outputs. This can make the training very difficult. Here, noting that both the stress and the strain are periodic in the $x$ direction, we choose to fit the stress by column. We train an LSTM to predict one column of the stress (the stress at the same $x$ in the physical space), using strains at this column and the neighboring columns. Figure \ref{fig:fitcol} shows this idea. In practice, when predicting the $k$-th column of the stress, we use strains from the $(k-2)$-th to the $(k+2)$-th column.

For the $2$-dimensional shear flow problem, we only show results from the direct training model. The LSTM is trained by an Adam optimizer for $10^5$ steps, with batch size being 64. Still, the stress is predicted using the strains from the last $20$ time steps.

\begin{figure}
\centering
\includegraphics[width=0.8\textwidth, height=0.35\textwidth]{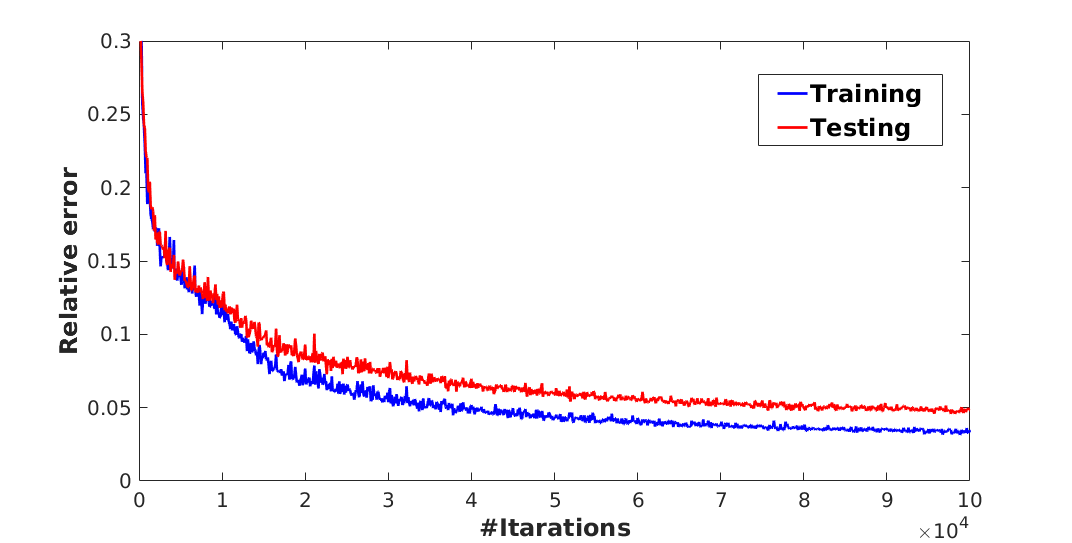}
\caption{Training and testing error of 2D shear flow.}\label{fig:2dtrain}
\end{figure}

\begin{figure}
\centering
\includegraphics[width=0.8\textwidth, height=0.35\textwidth]{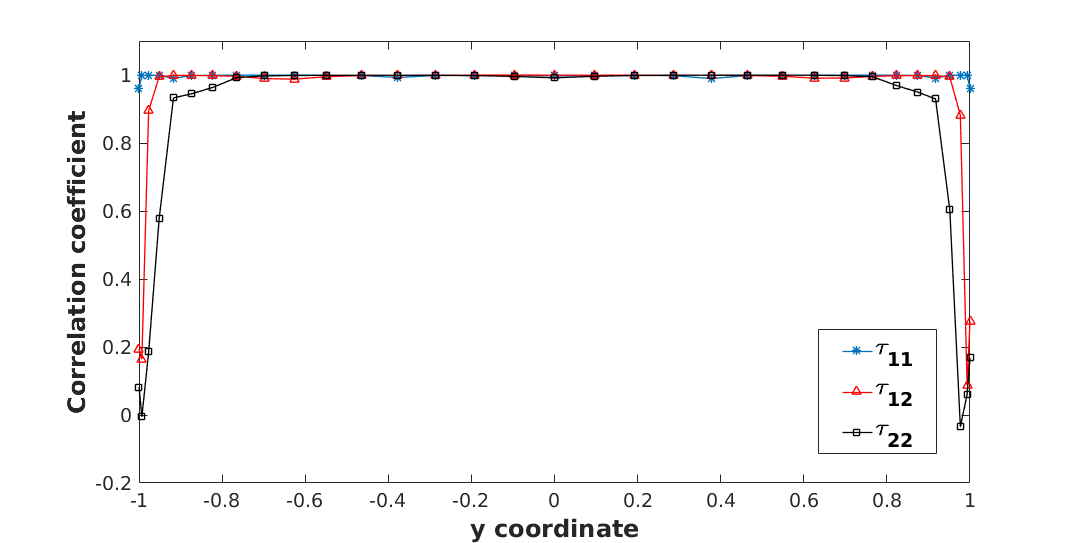}
\caption{ Correlation coefficients of predicted stress and true stress at different $y$.}\label{fig:corr}
\end{figure}

\paragraph{Numerical results}
Figure \ref{fig:2dtrain} shows the relative training and testing error during the training process. We can see that the training error goes down to $3-4\%$, while the testing error goes down to about $5\%$. 
Figure \ref{fig:corr} shows the correlation coefficients of the predicted stress and the true stress at different $y$. The three curves represent three components of the stress, respectively. we can see that the correlation coefficients are close to $1$ except in the area close to the boundary. Considering the boundary condition, the stress near the boundary is close to $0$. Hence, its reasonable for the prediction to have a low correlation with the true stress. 

\begin{figure}
\centering
\includegraphics[width=0.8\textwidth, height=0.35\textwidth]{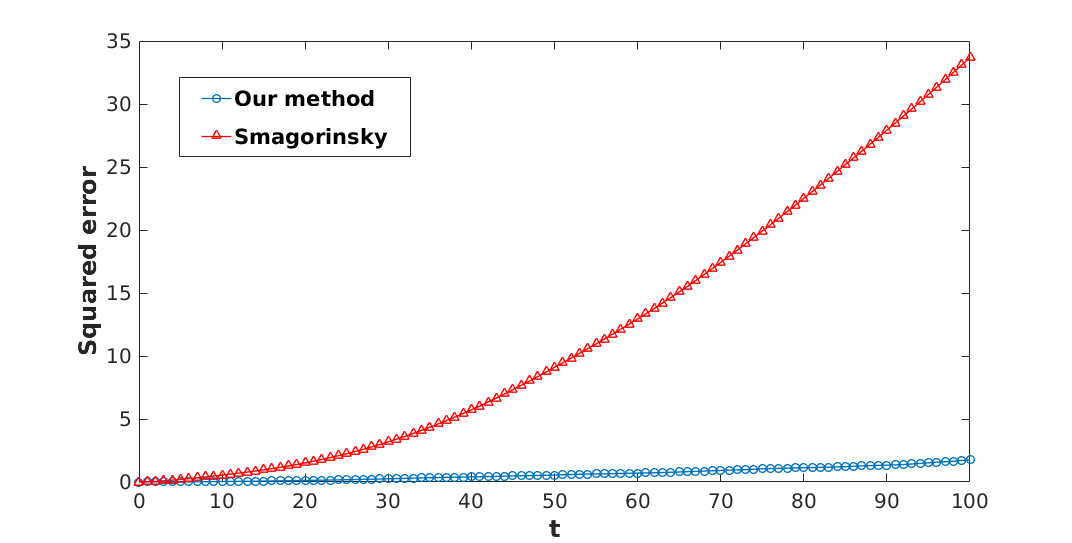}
\caption{Distance of the reduced solution and the true solution, compared with the Smagorinsky method.}\label{fig:error}
\end{figure}

Next, we solve the reduced system initialized from a true solution, and compare the quality of the reduced solution at later times with the solution given by the Smagorinsky model \cite{sm63}. The Smagorinsky model is a classical and widely used model for large eddy simulation (LES). In the two dimensional case, the Smagorinsky model for the sub-grid stress can be written as
\begin{equation}
\tau_{ij}-\frac{1}{2}\tau_{kk}\delta_{ij}=-2\nu_t\bar{S}_{ij},
\end{equation} 
where $\tau_{kk}=\tau_{11}+\tau_{22}$, $\nu_t$ is called the turbulent eddy viscosity, and
\begin{equation*}
\bar{S}_{11}=\frac{\partial \bar{u}}{\partial x},\ \bar{S}_{12}=\bar{S}_{21}=\frac{1}{2}\left(\frac{\partial\bar{u}}{\partial y}+\frac{\partial\bar{v}}{\partial x}\right),\ \bar{S}_{22}=\frac{\partial \bar{v}}{\partial y}.
\end{equation*}
The turbulent eddy viscosity can be expressed as 
\begin{equation}
\nu_t=(C_s\Delta)^2\sqrt{2\bar{S}_{ij}\bar{S}_{ij}},
\end{equation}
with $\Delta$ being the grid size and $C_s$ being a constant. 
In Figure \ref{fig:error}, we compare the deviation of our reduced solution and Smagorinsky solution from the true solution.
We see that the prediction given by our reduced system is much better than that by the Smagorinsky model. 

\section{Conclusion}\label{sec:con}

Much work needs to be done to develop the methods proposed here  into a systematic and practical approach for model reduction for a wide
variety of problems. For example, in the coupled training model, it is crucial to find an accurate and efficient way to compute the Jacobian. How to make the method scalable for larger systems is a problem that should be studied. 
For reduced systems where the noise effects cannot be neglected, how to model the noise term in the GLE is a problem that
should be dealt with.  Finally, we also need theoretical understanding to answer questions such as how to choose the size of the neural network, and how large the dataset should be in order  to obtain a good reduced system.

\bibliographystyle{unsrt}
\bibliography{01}

\end{document}